\documentclass[conference]{IEEEtran}
\IEEEoverridecommandlockouts

\usepackage{cite}
\usepackage{amsmath,amssymb,amsfonts}
\usepackage{algorithm,algorithmic}
\usepackage{graphicx}
\usepackage{textcomp}
\usepackage{xcolor}
\usepackage{booktabs}
\usepackage{multicol}
\usepackage{multirow}
\usepackage{url}
\usepackage{xcolor}

\def\BibTeX{{\rm B\kern-.05em{\sc i\kern-.025em b}\kern-.08em
    T\kern-.1667em\lower.7ex\hbox{E}\kern-.125emX}}
\begin{document}

\title{Improved Extrinsic Calibration of Acoustic \\Cameras via Batch Optimization}

\author{Zhi Li, Jiang Wang, Xiaoyang Li, and He Kong
\thanks{Zhi Li and Jiang Wang contributed equally to this work. This work was supported by the Science, Technology, and Innovation Commission of 
Shenzhen Municipality, China, under Grant No. ZDSYS20220330161800001, the Shenzhen Science and Technology Program under Grant No. KQTD20221101093557010, the Guangdong Science and Technology Program (Grant No. 2024B1212010002), and the National Natural Science Foundation of China 
under Grant No. 62350055. The authors are with the Shenzhen Key Laboratory of Control Theory and Intelligent Systems, and the Guangdong Provincial Key Laboratory of Fully Actuated System Control Theory and Technology, at the Southern University of Science and Technology, Shenzhen 518055, China. Emails:[12132272,12132297,12112611]@mail.sustech.edu.cn; kongh@sustech.edu.cn.}
	}

\maketitle

\begin{abstract}
Acoustic cameras have found many applications in practice. Accurate and reliable extrinsic calibration of the microphone array and visual sensors within acoustic cameras is crucial for fusing visual and auditory measurements. Existing calibration methods either require prior knowledge of the microphone array geometry or rely on grid search which suffers from slow iteration speed or poor convergence. To overcome these limitations, in this paper, we propose an automatic calibration technique using a calibration board with both visual and acoustic markers to identify each microphone position in the camera frame. We formulate the extrinsic calibration problem (between microphones and the visual sensor) as a nonlinear least squares problem and employ a batch optimization strategy to solve the associated problem. Extensive numerical simulations and real-world experiments show that the proposed method improves both the accuracy and robustness of extrinsic parameter calibration for acoustic cameras, in comparison to existing methods. To benefit the community, we open-source all the codes and data at 
\url{https://github.com/AISLAB-sustech/AcousticCamera.}

\end{abstract}


\section{Introduction}
Acoustic cameras are crucial for fusing visual and auditory perception in many applications, including industrial inspection \cite{Hyunuk2012,fiebig2020use,inoue2016three}, environmental monitoring\cite{AcousticFusion, Qian2018, negahdaripour2007epipolar, Jonas2020, Verellen2020, fu2024},
navigation\cite{Schouten2019, Steckel2013}
and underwater target detection \cite{wang2020acoustic,artero2021high,Yang2020,wang2020ral}, etc. The performance of acoustic cameras heavily depends on the accurate calibration of their parameters. Geometric parameters of acoustic cameras can be manually obtained for systems with pre-assigned structures \cite{allen2002aeroacoustic}. However, such procedures become infeasible when the acoustic sensor array structure becomes more complex. One can also use more specialized equipment such as Faro arms or laser scanners \cite{birchfield2003geometric,birchfield2005microphone}. Nevertheless, these equipment are expensive and might not be practical in many applications.

\begin{figure}[t]   
    \centering
   \includegraphics[width=0.7\columnwidth]{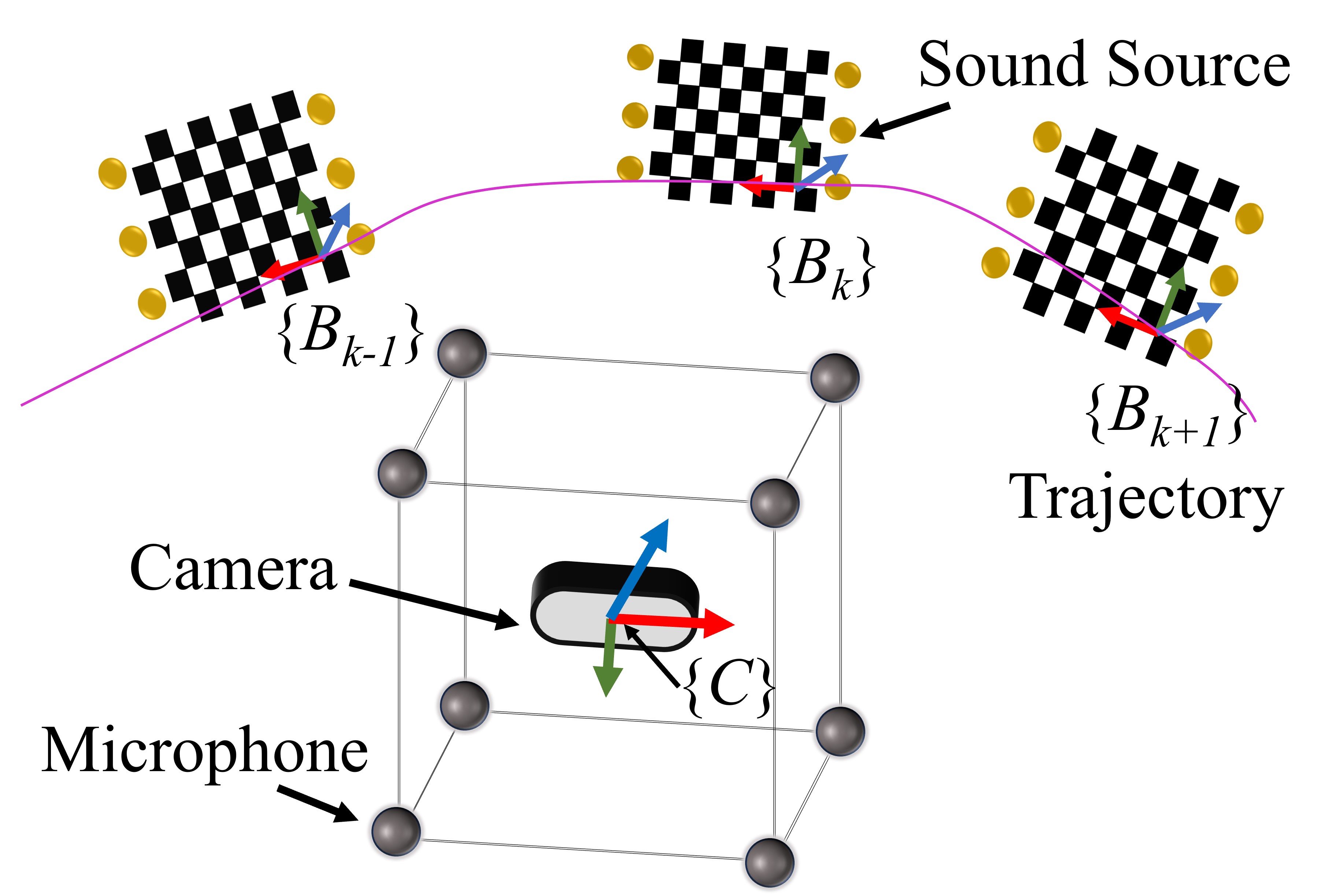}    
\caption{Extrinsic calibration of an acoustic camera using a moving calibration board with both visual and
acoustic marks}
\label{Geometry}
\end{figure}

In this paper, we consider acoustic cameras comprised of a microphone array and an optical sensor, and focus on estimation of relative positions (i.e., the acoustic camera extrinsic parameters) between the optical sensor and microphones \cite{O'Donovan, Legg2013}. Note that a substantial body of literature exists on microphone array calibration \cite{su2021necessary,sekiguchi2016online,Wang2024,HU2021,Plinge2016,zhang2024asynchronous,li2024} and camera calibration \cite{zhangcali,aprilcal,Shen2018,roland19}. Automatic calibration methods for audio-visual systems, however, has received much less attention \cite{Legg2013,yang2011development}.

More specifically, in \cite{yang2011development}, a sound target is used to align sound localization coordinates with image pixels to estimate the relative pose between the camera frame and the microphone array frame. However, this approach requires prior knowledge of the microphone array geometry.
A calibration method of grid search style is presented in \cite{Legg2013}. Similar to \cite{yang2011development}, the method in \cite{Legg2013} relies on rough information of the array geometry to pre-estimate the relative pose between the camera and the microphone array; afterward, a grid search is conducted for estimating each microphone position in the array frame and refine the relative pose estimates between the microphone array and the camera. Note that to fuse visual and auditory measurements of acoustic cameras, knowing the microphone positions in the camera frame is enough. In other words, there is no need to estimate the relative pose between the microphone array and the camera.

Motivated by the above observation, in this paper, we propose a novel and streamlined approach for automated extrinsic calibration of the acoustic camera using time difference of arrival (TDOA) acoustic measurements and a moving calibration board with both visual and acoustic marks, as shown in Fig. 1.
Instead of estimating the relative transformation between the
microphone array and the camera as in \cite{yang2011development, Legg2013}, we directly identify each microphone position in the camera frame. This not only helps to avoid requiring the prior knowledge of array geometry needed in \cite{yang2011development, Legg2013} but also allows us to formulate the calibration problem as a nonlinear least squares (NLS) optimization problem and solve it more effectively (than grid search in \cite{Legg2013}) using batch optimization. Extensive simulations and real-world experiments demonstrate that our method outperforms the calibration method in \cite{Legg2013}.

\section{The Proposed Method}
\label{sec:method}
\subsection{Problem Formulation}
As depicted in Fig. 1, we consider a typical scenario for extrinsic calibration of an acoustic camera with $N$ fixed microphones, an optical sensor,  and a moving calibration board with $M$ sound sources at known positions (within the board). The global frame is chosen to coincide with the camera frame $\{C\}$.
The position of the 
$i\raisebox{0mm}{-}th$ microphone in the global frame is denoted as $\mathbf{x}_{mic\_i}$, where $i=1,2,\cdots,N$. 

In the calibration process, we place the calibration board at different angles, and the camera captures image $P_k$ of the checkerboard at different positions, where $k=1,\cdots, K$. Simultaneously, at each checkerboard position, the $M$ sound sources emit signals sequentially, which are captured by the microphone array. 

Based on detection the pixel locations of the feature points at the corners of the checkerboard pattern, one can adopt existing methods (see \cite{zhangcali} and the references therein) to accurately solve the rotation matrix $\mathbf{R}_k$ from the global frame to the checkerboard frame $\{B_k\}$, as well as the position $\mathbf{t}_{k}$ of the checkerboard in the global frame. Moreover, the sound source position  $\mathbf{s}_{k,j}$ in the global frame, $j=1,\cdots,M$, can be expressed as follows:
\begin{equation}
\label{rw2cam}
    {\mathbf{s}}_{k,j} = \mathbf{R}_k {\mathbf{s}}_j^{B} + \mathbf{t}_k,
\end{equation}
where ${\mathbf{s}}_j^{B}$ represents the $j\raisebox{0mm}{-}th$ sound source position on the calibration board.

Without loss of generality, we use microphone 1 as the reference microphone. When the sound source $j$ sends the acoustic signal at position $\mathbf{s}_{k,j}$, the analytical expression of TDOA measurement between the $i$-$th$ microphone and the reference microphone can be obtained as:
\begin{equation}
    T^k_{i1} = \frac{|| \mathbf{x}_{mic\_i} - \mathbf{s}_{k,j} ||}{c}-\frac{|| \mathbf{x}_{mic\_1} - \mathbf{s}_{k,j} ||}{c}
    \label{tdoa}
\end{equation}
for $i=2,\cdots,N$, $j=1,\cdots,M$, where $c$ represents the sound speed in the air. Thus, the whole unknown  extrinsic parameters (i.e., the position of each microphone in the camera frame) to be calibrated in this paper can be denoted as:
\begin{equation}
\mathbf{x}= \left [ \mathbf{x}_{mic \_ 1};\mathbf{x}_{mic \_ 2};\cdots;\mathbf{x}_{mic \_N} \right ].
\end{equation}

Denote the ideal TDOA measurements at the $k$-$th$ calibration board position as:
\begin{equation}
    \mathbf{m}_k = \left[ T_{21}^k ; T_{31}^k ; \cdots; T_{N1}^k \right ] \in  \mathcal{R} ^{N-1}.\label{m}
\end{equation}
Assume TDOA measurements in (\ref{m}) are affected by Gaussian noise, as follows:
\begin{equation}
    \mathbf{y}_k = \mathbf{m}_k + \mathbf{v}_k, 
\end{equation}
where $\mathbf{v}_k \sim \mathcal{N}(\mathbf{0},\mathbf{P}) $, with 
$\mathbf{P}>\mathbf{0} \in \mathcal{R}^{(N-1)\times(N-1)}$. All measurements of the microphone array can be represented as:
\begin{equation}
\begin{aligned}\mathbf{z} & =\left[\mathbf{y}_1;\mathbf{y}_2;\cdots;\mathbf{y}_K\right]\\
& =\mathbf{g(x)}+ \mathbf{\gamma},
\end{aligned}
\label{eq_Z}
\end{equation}
where $\mathbf{g(x)}$ is the observation model, and $\gamma \sim \mathcal{N} (\mathbf{0},\mathbf{W})$ is the noise of observation with
\begin{equation}
    \mathbf{W} = diag_{K}(\mathbf{P}).
\end{equation}
Then, following the ideas and concepts in \cite{su2021necessary}, the problem of calibrating the extrinsic parameters of an acoustic camera can be treated as the following NLS problem:
\begin{equation}
    \underset{\mathbf{x}} {\mathrm{min} }\left \| \mathbf{g}(\mathbf{x}) -\mathbf{z} \right \|^2_{\mathbf{W} ^{-1}}
    \label{nls}
\end{equation}
The measurements obtained by the microphone array and
the sound sources constitute spatial constraints and
can be included in (\ref{nls}) to improve the accuracy of the estimation.

Given the context provided above, our primary objective is to achieve a calibration of the extrinsic parameters of the acoustic camera using a calibration board with both visual and acoustic markers.

\subsection{Batch Optimization}

As a general NLS problem, (\ref{nls}) does not have a closed-form solution and requires numerical optimization techniques to iteratively solve for the solution \cite{maye2016online}. To illustrate, in this paper, we employ the Gauss-Newton iterations to solve the corresponding NLS problem. Other improved versions of Gauss-Newton iterations or more advanced nonlinear solvers may be adopted for better convergence and performance. However, due to limited space, we will not elaborate on the details.

The increment $\mathbf{\Delta x}$ at each iteration and the estimation value $\hat{\mathbf{x}}$ at the $t\raisebox{0mm}{-}th$ iteration can be obtained by solving the following equation:
\begin{equation}
\begin{array}{l}
\label{Increment equation}
\mathbf{H}\mathbf{\Delta x} = - \mathbf{b},
\\
\hat{\mathbf{x}}_{t+1}=\hat{\mathbf{x}}_{t}+\Delta\mathbf{x},
\end{array}
\end{equation}
where $\mathbf{H}$ is the approximation matrix of the Hessian matrix and $\mathbf{b}$ is the coefficient vector:
\begin{equation}
\mathbf{H}=\mathbf{J}^{\top}\mathbf{W}^{-1}\mathbf{J},\text{ }\mathbf{b}=\mathbf{J}^{\top}\mathbf{W}^{-1}\mathbf{e},\label{eq_LS}
\end{equation}
where the vector $\mathbf{e}$ is a difference between the expected observation $\mathbf{g}(\hat{\mathbf{x}}_t)$ and the real observation $\mathbf{z}$ gathered by the microphones;
the matrix $\mathbf{J}$ is the Jacobian matrix of the error function $\mathbf{g}(\mathbf{x})-\mathbf{z}$ with respect to unknown parameters vector $\mathbf{x}$ and calculated at the value of $\hat{\mathbf{x}}_t$. Combining the camera calibration stage and the Gauss-Newton iteration procedure, we then have the entire acoustic camera calibration algorithm, as shown in Algorithm 1.

\begin{algorithm}[t]
\caption{Calibration of Acoustic Camera}
\label{alg} 
\begin{algorithmic} 
\REQUIRE Sensors measurements $\mathbf{z}$ 
\ENSURE Estimation of all unknown parameters $\widehat{\mathbf{x}}$

\STATE// Camera Calibration

Obtain the extrinsic parameters ($\mathbf{R}_k$ and $\mathbf{t}_k)$ of the camera for each image;
\FOR{$k\in[1,K]$}
\FOR{$j\in[1,M]$}
\STATE Compute the sound source positions $\mathbf{s}_{k,j}$ in frame $\{C\}$ with \eqref{rw2cam};
\ENDFOR
\ENDFOR
\STATE //Acoustic Camera Calibration

Update $\mathbf{x}_{mic\_i}$ for each microphone in turn until a minimum of \eqref{nls} is reached using Gauss-Newton iterations.
\end{algorithmic} 
\end{algorithm}

\section{Numerical Simulations and Results}
\label{sec:pagestyle}
In the simulation, an acoustic camera consisting of a microphone array and a camera is placed stationarily. The microphone array has eight microphones positioned at the vertices of a cube with a side length of 0.5 meters. The camera is placed at the cube's center, and the camera frame is regarded as the global frame. The parameters used in our simulations are listed in Table I. 

During the numerical simulations, the TDOA measurement errors were set to 0.0666$ms$, 0.333$ms$, 0.999$ms$ and 1.332$ms$, respectively. These scenarios are referred to as Lv 1, Lv 2, Lv 3, and
Lv 4, respectively. For each set of experiments, 100 rounds of Monte Carlo simulations were performed, with the initial positions of the microphone array randomly selected within the range of 0 to 0.5 meters. The root mean square error (RMSE) was then calculated to quantify the errors.

First, we evaluated the impact of reference microphone selection on the calibration process. In the setup denoted as Ours-1, a specified microphone was designated as the reference microphone, with its position considered unknown. As shown in \eqref{m}, this configuration resulted in 7 measurements at each calibration board position. In the experimental group labeled Ours-2, each microphone in the array was sequentially set as the reference microphone at each calibration board position, leading to
$N(N-1)/2$ TDOA measurements. The experimental results are presented in Table II under Exp 1.

Subsequently, we compared the proposed method with the method presented in \cite{Legg2013}. Note that the method in \cite{Legg2013} relies on a known position of the reference microphone relative to the initial calibration board frame $\{B_1\}$. To accommodate this, we added an additional reference microphone with a known position in frame $\{B_1\}$, resulting in a total of 9 microphones. The experimental group for the proposed method in this scenario is denoted as Our-3. The calibration results are presented in Table II under Exp 2.

It is evident that as the TDOA measurement noise increases, the final estimation error also gradually increases. In the simulation data, it is assumed that each microphone experiences the same level of noise. As a result, in Exp 1, there is little difference between the configuration where each microphone is sequentially set as the reference and the configuration with a single reference microphone. In Exp 2, our calibration results significantly outperform the method in \cite{Legg2013}, especially at lower noise levels. This is because \cite{Legg2013} relies on search-based optimization within a certain range to estimate the microphone positions, meaning that the size of the search region directly affects the final calibration accuracy.
\begin{table}[t]
\begin{center}
\caption{Simulation Parameters}
\begin{tabular}{p{5.5cm} p{1.8cm}}
\toprule
\textbf{Parameters} & \textbf{Values} \\
\midrule
Initial microphone position & 0-0.5m  \\
Sound source position error STD & 0.1m\\
Sound speed in air & 340m/s\\
Guass-Newton max iteration & 50 \\
Gauss-Newton iteration threshold & 0.001 \\
\bottomrule
\label{simu_para}
\end{tabular}
\vspace{-0.5cm}
\end{center}
\end{table}

\begin{table}[t]
\begin{center}
\renewcommand\arraystretch{1.4}
\caption{\label{tab:1}THE RMSE OF CALIBRATION RESULTS UNDER VARYING NOISE LEVELS: ANALYSIS OF 100 MONTE CARLO SIMULATIONS $\left(m\right)$ (Bold means better)}
\centering 
\begin{tabular}{c|cc|cc}
\hline
\multirow{2}{*}{TDOA} &	\multicolumn{2}{c|}{Exp 1} &	\multicolumn{2}{c}{Exp 2} \\
Noise& Ours-1 & Ours-2 & Ours-3 & \cite{Legg2013}   \\
\hline
Lv 1 & 8.136e-03  & \textbf{7.936e-03}  & \textbf{1.160e-02}  &  3.358e-01 \tabularnewline

Lv 2   & 4.290e-02& \textbf{4.203e-02} & \textbf{5.771e-02}   & 3.445e-01 \tabularnewline

Lv 3  & \textbf{1.438e-01} & 1.452e-01 & \textbf{1.747e-01}    & 4.682e-01 \tabularnewline

Lv 4   & 2.038e-01 & \textbf{1.939e-01} & \textbf{2.331e-01}   & 5.013e-01 
\tabularnewline
\hline
\end{tabular}
\end{center}
\end{table}

\section{Experiments and Results}

\subsection{Experimental Setup}

\begin{figure}[t]

\begin{minipage}[b]{0.48\linewidth}
  \centering
  \centerline{\includegraphics[width=0.97\linewidth]{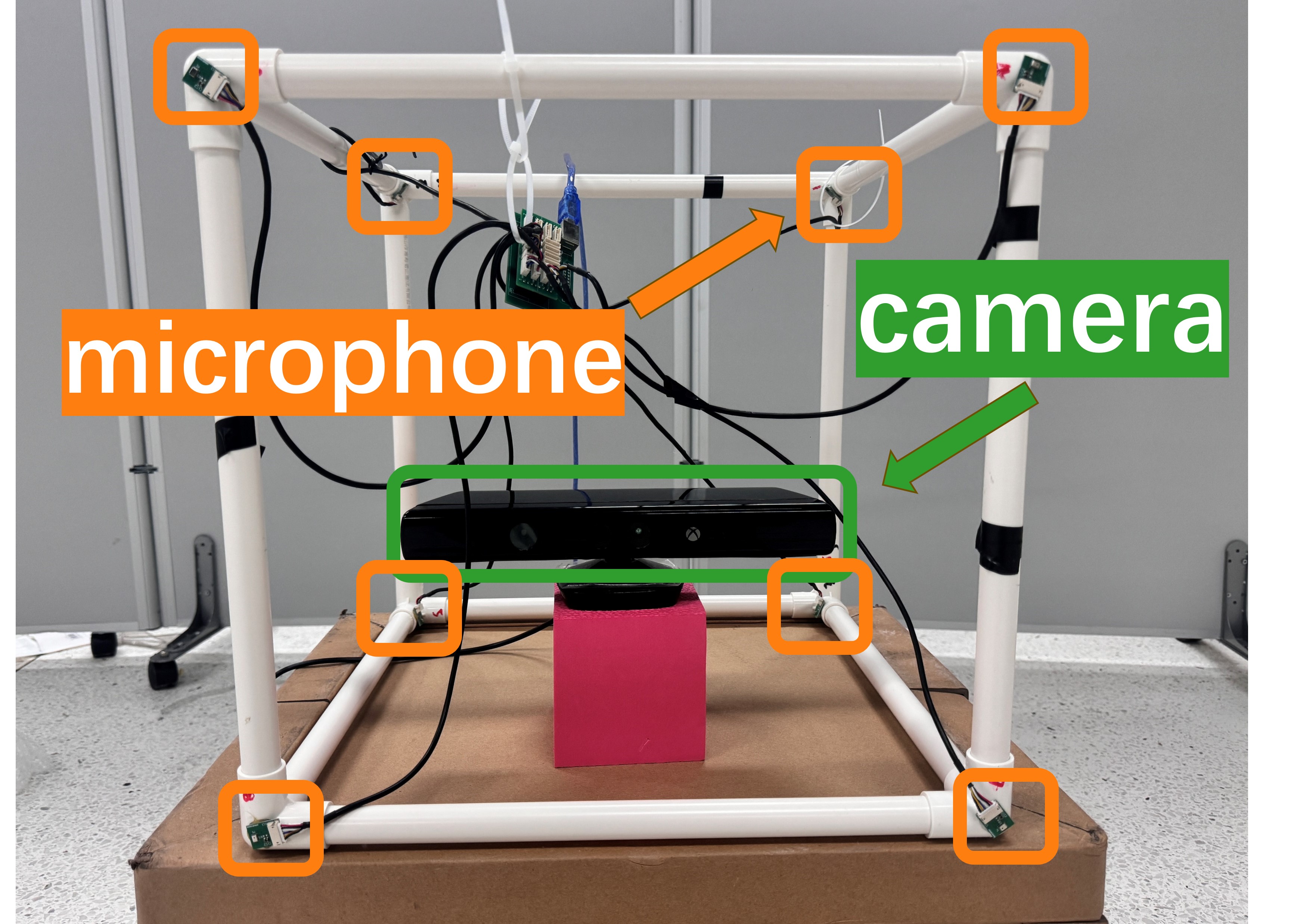}}
  \centerline{(a)}\medskip
\end{minipage}
\hfill
\begin{minipage}[b]{0.48\linewidth}
  \centering
  \centerline{\includegraphics[width=0.95\linewidth]{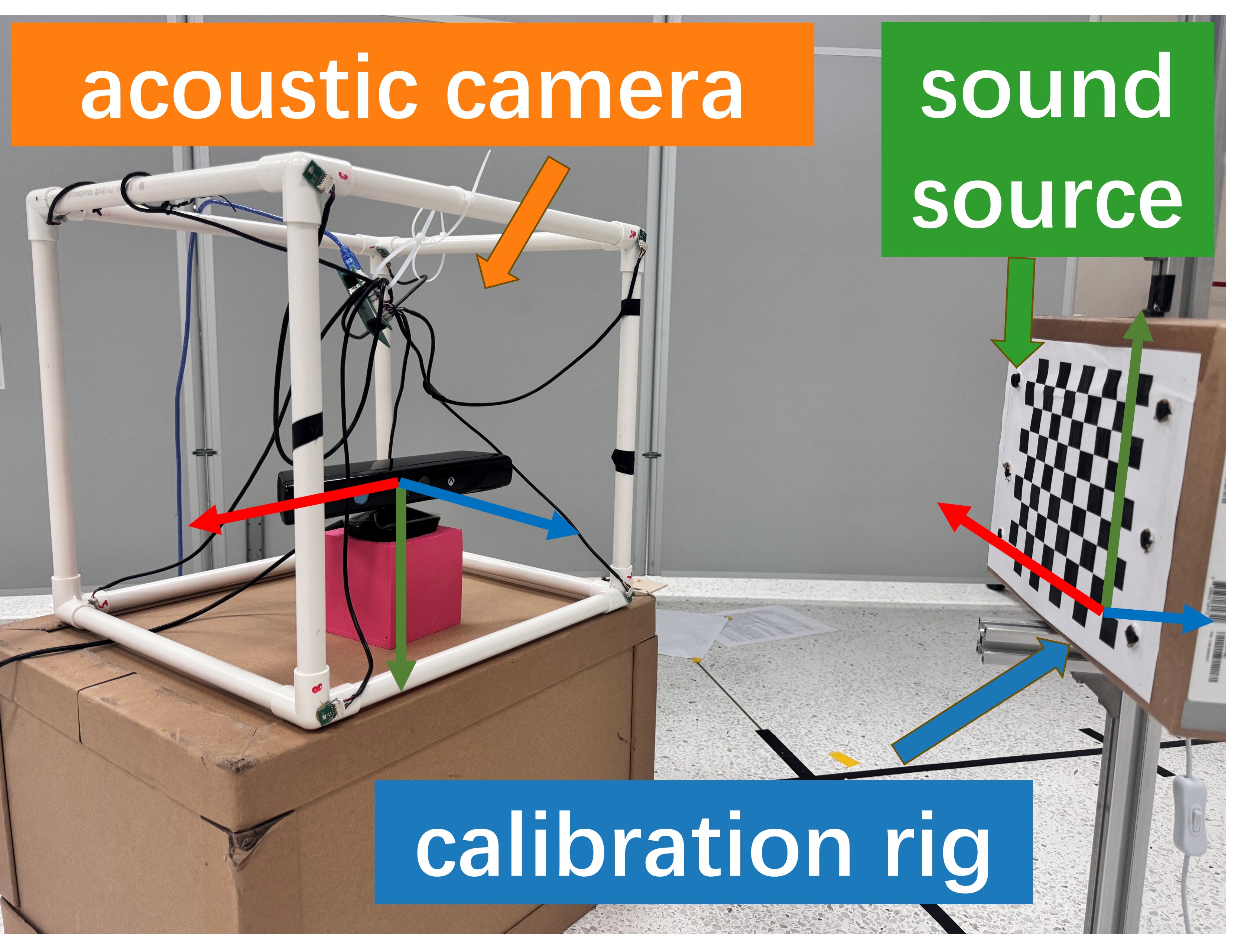}}
  \centerline{(b)}\medskip
\end{minipage}
\caption{Real-world 3D acoustic camera calibration environment setup. (a) Microphone and camera. (b) Typical calibration scenario.}
\label{micro&cam}
\end{figure}
Similar to the simulation setup, the acoustic camera to be calibrated consists of an 8-channel microphone array and a Kinect V1 camera, as shown in Fig. 2(a). Each microphone is positioned at the vertices of a cube with a side length of 0.4 meters. The calibration board, depicted in Fig. 2(b), includes a checkerboard pattern and six buzzers at known locations.

During the calibration process, the acoustic camera remains stationary while the calibration board is randomly moved to different positions within the camera’s field of view. At each position, the six sound sources emit signals sequentially, which are captured by the microphone array. By analyzing the checkerboard images captured from different angles and detecting the pixel locations of the checkerboard corners, the pose of the calibration board at each position can be accurately determined.

To measure the TDOA between the eight microphone channels, we employed the widely used generalized cross-correlation phase transform (GCC-PHAT)  algorithm \cite{knapp1976generalized}. In the final accuracy assessment of the acoustic camera extrinsic calibration, the ground truth position of each microphone was determined through direct ruler measurements. RMSE was then calculated to quantify the errors.

In the experiment, 69 images and 414 sound signals were collected. Using this dataset, similar to the simulation, we divided the experiments into four groups, Ours-1, Ours-2, Ours-3, and \cite{Legg2013}, to investigate the impact of different initial values on the proposed method and the \cite{Legg2013}. 

The initial microphone positions were randomly selected within 0.5m around the origin. Subsequently, to investigate the impact of varying dataset sizes on the calibration results, we conducted experiments by randomly selecting different numbers of images and corresponding sound signals from the dataset. For each scenario, 100 Monte Carlo experiments were conducted.

\begin{figure}[t]
\begin{minipage}[b]{0.9\linewidth}
  \centering
\centerline{\includegraphics[width=0.85\linewidth]{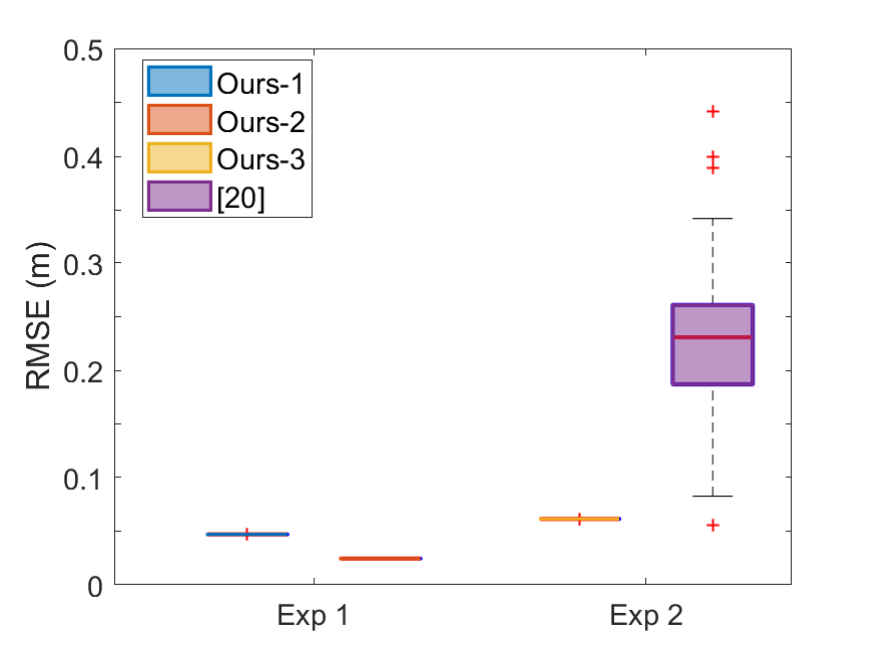}}
  \centerline{(a)}\medskip
\end{minipage}
\vspace{-0.5cm}
\begin{minipage}[b]{0.89\linewidth}
  \centering
\centerline{\includegraphics[width=0.86\linewidth]{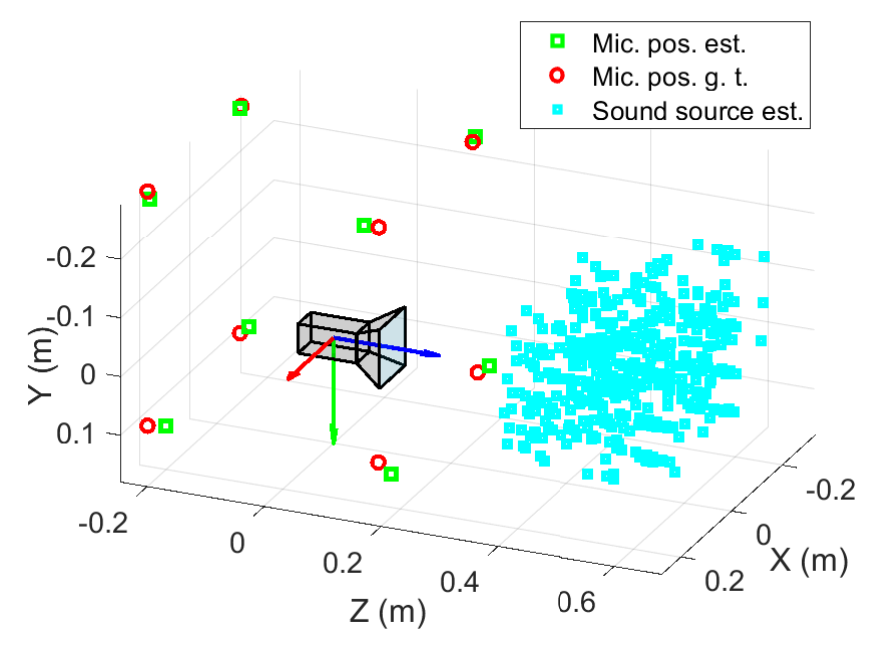}}
  \centerline{(b)}\medskip
\end{minipage}
\caption{Experiment results in the real-world experiment: 
(a) Box plot of microphone position estimation errors. 
(b) Calibration results and the true values of the acoustic camera.
}
\label{result}

\end{figure}
\subsection{Results and Discussion}
The calibration results are presented in Table \ref{tab:exp}. As shown, in real-world scenarios, sequentially designating each microphone as the reference microphone significantly improves the accuracy of the calibration results. This is because each microphone has different levels of measurement noise, and using each one as a reference, in turn, mitigates the impact of noise from any single reference microphone. Additionally, the results show that varying the initial microphone position error across three levels has little effect on the calibration outcomes.

In contrast, the method in \cite{Legg2013} relies on search-based optimization within a specific range to estimate the positions. 

As shown in Fig. \ref{result}(a), our proposed method produces fewer outliers (red cross mark) and higher accuracy for extrinsic calibration of the audio-visual system. Fig. \ref{result}(b) displays the final calibration result of the acoustic camera. 

\begin{figure}[t]
  \centering
  \includegraphics[width=0.95\linewidth]{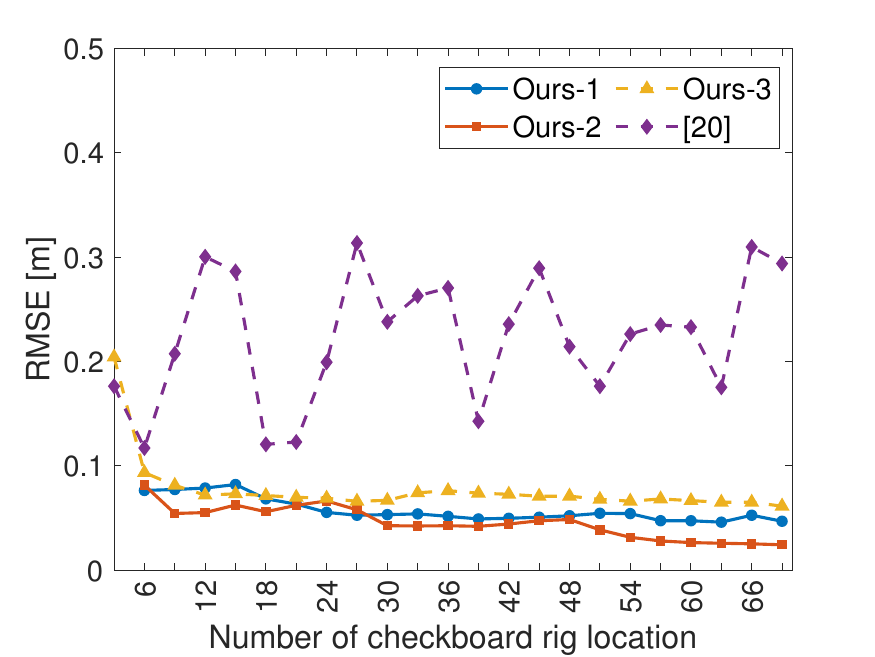}
\caption{Calibration results across different size of datasets.}
\label{result2}
\end{figure}

As shown in Fig. \ref{result2}, the calibration accuracy of the proposed method improves with an increasing number of measurements, while \cite{Legg2013} shows no significant change. This is because the proposed method benefits from additional measurements by incorporating more constraints and providing information from multiple perspectives. In contrast, the method in \cite{Legg2013} has a grid search mechanism, making it prone to converge to local optima, limiting its potential to improve accuracy as the number of measurements increases.
\begin{table}[t]
\begin{center}
\renewcommand\arraystretch{1.4}
\caption{\label{tab:exp}THE RMSE OF CALIBRATION RESULTS UNDER DIFFERENT INITIAL VALUES $\left(m\right)$ (Bold means better)}
\begin{tabular}{cc|cc}
\hline
\multicolumn{2}{c|}{Exp 1} &	\multicolumn{2}{c}{Exp 2} \\
Ours-1 & Ours-2 & Ours-3 &  \cite{Legg2013}    \\
\hline
4.701e-02  &  \textbf{2.444e-02}  & \textbf{6.146e-02}  &  9.129e-02
\tabularnewline

\hline
\end{tabular}
\end{center}
\end{table}
\label{sec:typestyle}

\section{Conclusions}
\label{sec:majhead}

In this paper, we propose an effective batch optimization-based automatic calibration technique for acoustic cameras. By using a calibration board equipped with both visual and acoustic markers, we formulate the extrinsic calibration problem (between microphones and the optical camera) as an NLS program and employ a batch optimization strategy to solve the associated problem. As demonstrated by extensive numerical simulations and real-world experiments, the proposed method renders improved accuracy and robustness for extrinsic parameter calibration, compared to existing methods.

\clearpage
\bibliographystyle{IEEEtran}
\bibliography{conference}
\end{document}